\DocumentMetadata{}

\documentclass[sigconf]{acmart}

\AtBeginDocument{%
  }

\copyrightyear{ }
\acmYear{}
\setcopyright{acmlicensed}\acmConference[GECCO '24]{Genetic and Evolutionary Computation Conference}{July 14--18, 2024}{Melbourne, VIC, Australia}

\usepackage{upgreek}
\usepackage{algorithm}
\usepackage{amsmath}
\usepackage{booktabs}
\usepackage{enumerate}
\usepackage{multirow}
\usepackage{makecell}
\usepackage{algpseudocode}
\usepackage{graphicx}
\usepackage{caption}
\usepackage{subcaption}
\usepackage{url}
\usepackage{todonotes}
\usepackage{float}

\begin{document}

\title{Promoting Two-sided Fairness in Dynamic Vehicle Routing Problem}

\author{Yufan Kang}

\email{yufan.kang@student.rmit.edu.au}
\affiliation{%
  \institution{RMIT University}
  \city{Melbourne}
  \state{VIC}
  \country{Australia}
}

\author{Rongsheng Zhang}
\email{zhang.rongsheng@hotmail.com}
\affiliation{%
  \institution{RMIT University}
  \city{Melbourne}
  \state{VIC}
  \country{Australia}
  }

\author{Wei Shao}
\email{wei.shao@data61.csiro.au}
\affiliation{%
  \institution{Data61, CSIRO}
  \city{Clayton}
  \state{VIC}
  \country{Australia}
}

\author{Flora D. Salim}
\email{flora.salim@unsw.edu.au}
\affiliation{%
 \institution{University of New South Wales}
 \city{Sydney}
 \state{NSW}
 \country{Australia}}

\author{Jeffrey Chan}
\email{jeffrey.chan@rmit.edu.au}
\affiliation{%
\institution{RMIT University}
  \city{Melbourne}
  \state{VIC}
  \country{Australia}
}

\renewcommand{\shortauthors}{Kang et al.}

\begin{abstract}
    Dynamic Vehicle Routing Problem (DVRP), is an extension of the classic Vehicle Routing Problem (VRP), which is a fundamental problem in logistics and transportation. Typically, DVRPs involve two stakeholders: service providers that deliver services to customers and customers who raise requests from different locations. Many real-world applications can be formulated as DVRP such as ridesharing and non-compliance capture. Apart from original objectives like optimising total utility or efficiency, DVRP should also consider fairness for all parties.  Unfairness can induce service providers and customers to give up on the systems, leading to negative financial and social impacts. However, most existing DVRP-related applications focus on improving fairness from a single side, and there have been few works considering two-sided fairness and utility optimisation concurrently. To this end, we propose a novel framework, a Two-sided Fairness-aware Genetic Algorithm (named 2FairGA), which expands the genetic algorithm from the original objective solely focusing on utility to multi-objectives that incorporate two-sided fairness. Subsequently, the impact of injecting two fairness definitions into the utility-focused model and the correlation between any pair of the three objectives are explored. Extensive experiments demonstrate the superiority of our proposed framework compared to the state-of-the-art.
\end{abstract}




\maketitle

\section{Introduction}
\label{Introduction}

The Dynamic Vehicle Routing Problem (DVRP) is a logistical challenge that focuses on managing the routes and schedules of multiple vehicles in a changing environment, with two primary stakeholders at its core: i) service providers, which include logistics companies, delivery services, or any organisation managing the vehicle fleet, and ii) customers, who raise requests at different locations and expect reliable service deliveries. In this paper, we focus on the dynamic aspect of customer requests, which means new service requests may arrive randomly over time, requiring incorporation into the existing route plan while optimising different objectives. Typically, balancing the needs of these stakeholders is crucial in DVRP; service providers strive to optimise operations while maintaining high levels of customer satisfaction~\cite{schmoll2021semi,tong2020spatial,seghezzi2021demand}, a task that becomes increasingly complex with real-time variables such as changing customer demands. Ridesharing, food delivery and stochastic resource allocation are all important examples of applications in DVRPs with multiple stakeholders/parties. These real-world applications include vehicles that travel to serve dynamically changing customer requests in real-time, and the constructed platforms need to schedule and optimise routes for the vehicles to balance two parties.  
Recent findings suggest that various proposed algorithms can enhance efficiency and utility across applications, thereby boosting service providers' profits and supporting the sustained growth of platforms \cite{han2023path,yao2021dynamic,tong2020spatial,wang2022recommending}. However, existing allocation systems for DVRPs inadvertently have negative consequences on fairness, such as unequal working load among service providers and waiting time gap among customers \cite{gupta2022fairfoody}. For example, some service providers might have to work longer while getting the same earnings as their job allocations are not profitable. Some customers in less serviced areas might have to wait longer \cite{suhr2019two}. These side effects are referred to as fairness issues, in which disadvantaged service providers get lower wages and disadvantaged customers get worse services.

It is imperative to design an allocation system that simultaneously considers two-sided fairness and utility.  Existing studies only consider single-sided fairness and further research is required to reveal the interrelationships among three objectives.  They focus on improving fairness among service providers while optimising utility. To consider fairness issues among service providers, prior studies have proposed models to strike a balance between improving fairness among service providers and utility optimisation \cite{raman2021data,Shi2021,Lesmana2019,Xu2020,Nanda2020,suhr2019two,gupta2022fairfoody} under real-world scenarios such as ridesharing and food delivery. These studies have also explored the correlation between fairness among service providers and utility, and revealed that there exists a negative correlation between these two objectives \cite{raman2021data,Shi2021,Lesmana2019,Xu2020,Nanda2020,suhr2019two,gupta2022fairfoody}. As for the fairness issues among customers, only few studies focus on improving fairness among customers while optimising utility \cite{suhr2019two,wu2021fairness,Brandao2020}. The relationship between customer fairness and optimising utility is unexplored and needs further analysis. In the realm of two-sided fairness and utility optimisation within a single system, although the study \cite{suhr2019two} proposed a method to explore fairness on both sides while optimising utility, the two-sided fairness is considered separately.

To consider two-sided fairness and optimising utility in a single model for DVRP, the challenge is mainly due to: i) increased chance of conflict between the objectives \cite{yu2021multi,pan2021practical,li2019predictive}; and ii) the decision-making complexity \cite{jain2013evolutionary,soleimani2015transmission,ma2022critical}. On the one hand, improving fairness from either side may induce conflict with the utility optimisation objective. For instance, to ensure fairness among customers, a vehicle might be routed to a farther destination to equalise waiting times, even if closer destinations are available. This could decrease the total utility gained. On the other hand, improving fairness among customers may increase the total utility gained. For instance, to ensure fairness among customers, the system may encourage prioritising certain critical destinations that may be consistently deprioritised in a utility-driven model. Additionally, as the system dynamics change with time, such as the introduction of new destinations or changes in traffic conditions, the decision-making process becomes even more intricate. The dynamic nature of the problem can amplify the potential for conflicts between two-sided fairness and utility, making it challenging to find solutions that satisfy all three objectives simultaneously.

Faced with these gaps, we take a further step by promoting two-sided fairness to the underlying systems. We propose an innovative Genetic Algorithm (GA) based approach that incorporates fairness on both sides (service providers and customers). We first consider formulating different real-world scenarios as DVRPs, including multiple service providers initialised at different locations and customers raising requests from different locations. Then, to achieve two-sided fairness, we propose a Two-sided Fairness-aware Genetic Algorithm (2FairGA) by replacing the two evaluation metrics from optimising utility to improving service provider-based fairness and improving customer-based fairness accordingly. To allow the proposed method to consider utility optimisation at the same time, we leave the rest of the steps in the Genetic Algorithm unchanged. Considering two-sided fairness and optimising utility in different steps allows the proposed method to alleviate the conflict between the three objectives in comparison to directly constructing a trade-off function using the objectives such as a linearly weighted sum. In addition, 2FairGA utilises constrained K-means clustering to initialise the locations of the service providers as a sampling process to enhance two-sided fairness further. To explore the generality of the proposed method, we use two real-world scenarios that can be formulated as DVRP and experiencing fairness issues on both sides: non-compliance capture and ridesharing, which shows that 2FairGA can adapt to different fairness definitions.

The main contributions of this paper are summarised as follows:
\begin{itemize}
\item Our work is the first to incorporate the concept of two-sided fairness and utility optimisation to solve Dynamic Vehicle Routing Problems such as the travelling officer problem and ridesharing problem.
\item To balance utility optimisation and two-sided fairness, we introduce two distinct fairness objectives and integrate them into a Genetic Algorithm.
\item We demonstrate the importance of sampling the drivers by using a clustering algorithm to set their starting points, which further enhanced two-sided fairness.
\item We conduct experiments on both a publicly available large-scale ridesharing dataset and a parking non-compliance dataset to validate the effectiveness of our proposed approach and explore the correlation between two-sided fairness and utility optimisation.
\end{itemize} 

\section{Related Work}
\label{related_work}
\textbf{Utility Optimisation in DVRP:} Over the past few decades, Dynamic Vehicle Routing Problems (DVRPs) have been widely studied. It is variated based on Vehicle Routing Problem, which NSGA-II and NSGA-III have been employed with high-performance allocation plans \cite{deb2002fast,deb2013evolutionary,geng2022dual}, but failed to capture dynamic features for DVRPs. DVRPs are characterised by its real-time and evolving nature~\cite{psaraftis2016dynamic}. In the case of utility optimisation, the main objective for DVRP is to either maximise the total utility gained through the entire time duration or minimise the time cost to complete all requests that happened in the time duration. Some meta-heuristic algorithms, such as genetic optimisation \cite{taniguchi2004intelligent}, have been applied to the problem. Apart from meta-heuristic algorithms, there exist studies that consider reinforcement learning-based methods to solve the problem. For instance, Ulmer \textit{et al.} proposed a route-based Markov Decision Process-based model for addressing DVRPs \cite{ulmer2017route}. In addition to single-objective DVRPs, Ghannadpour \textit{et al.} proposed a Multi-Objective Dynamic Vehicle Routing Problem (MODVRP) and tried to find the Pareto front for two objectives \cite{ghannadpour2013multiobjective}. An example of DVRP can be non-compliance capture problems, which are formulated as Multi-agent Travelling Officer Problem (MTOP). The main purpose for MTOP is to optimise utility which is defined to be non-compliance capture rate, and Qin \textit{et al.} proposed a Leader-based Random Keys Encoding Scheme (LERK) to encode sequences of officers and parking bays \cite{qin2020solving}. For non-compliance capture problems, previous studies have mainly focused on optimisation and have ignored fairness among officers and parking non-compliance. 

\textbf{Fairness in DVRP} Many studies have been conducted to address fairness and efficiency in Dynamic Vehicle Routing Problems. For example, in 2020, Gollapudi \textit{et al. } proposed a model focusing on envy-freeness for repeated matching in two-sided markets, and they argue that in an asymmetric (the number of customers and the number of service providers are different) dynamic environment, it is impossible to output a solution that guarantees envy-freeness up to a single match and selects a maximum weight matching on each time step, which shows the negative correlation between fairness and optimising utility ~\cite{Gollapudi2020}. Additionally, there are studies that focus on fairness issues in real-world applications, such as food delivery \cite{gupta2022fairfoody}, mobile robots navigation systems \cite{Brandao2020}, and ridesharing \cite{sun2022optimizing,raman2021data,Nanda2020,Xu2020,Lesmana2019}, that can be formulated as DVRPs. For ridesharing, in 2019, Suhr \textit{et al.} first presented a novel framework for reevaluating fairness in ride-hailing platforms, with an emphasis on driver well-being. The proposed framework for achieving fairness in ride-hailing platforms focuses on equitable income distribution for drivers over time \cite{suhr2019two}. In 2021 Raman \textit{et al.} stated that traditional optimisation algorithms can only provide myopic allocation plans for ridesharing, and they aimed to decrease inequality in ride-pooling platforms by incorporating fairness constraints into the objective function for Markov Decision Process (MDP) -based framework and redistributing income among drivers \cite{raman2021data}. 
Previous research has focused on improving fairness among drivers and overall utility, exploring their relationship but overlooking customer fairness and the integration of two-sided fairness with optimised utility within a single framework. The interplay among these three objectives remains insufficiently examined.

\section{Preliminary}
\label{preliminary}

\subsection{Problem Formulation}

We provide the problem formulation of DVRP. Let the real-world map over which a typical DVRP optimises over is modelled as a directed graph $G = (V, \mathcal{E})$, in which $V := \{v_1, v_2, v_3, ..., v_m\}$ is the set of all nodes (locations) and $\mathcal{E}:V \times V$ is the set of all edges (roads connecting different locations). A directed edge from $v_i$ to $v_j$ exists, e.g., $(v_i, v_j) \in \mathcal{E}$, if there is a road that allows a vehicle to travel from $v_i$ to $v_j$. Assume there is a set of service providers $M$ starting from different locations, and each driver $n \in N$ is characterised by a set of attributes $n := \{u_n, l_n\}$, where $u_n$ represents the accumulated utility that is acquired by $n$ across time, and $l_n$ is current location of a driver, which is one of the node locations, i.e, $l_n \in V$ represents the current location of $n$. Let there be a set of customers $C$ raising requests from different locations and at different time points, and each customer $c \in C$ is characterised by a set of attributes $c := \{s_c, d_c, T_c\}$, where $s_c\in V, d_c\in V$ represents the start and destination locations/nodes for the customer, and $T_c = (t^c_s, t^c_e)$ represents the time duration the current request is valid, with $t^c_s$ representing the starting time and $t^c_e$ represents the latest time when the request should be serviced.

In DVRP, the total utility acquired in time duration $T$ is defined as $\sum _{n\in N} u_n$. Though definition for service provider-based fairness is different in DVRP, we define service provider-based fairness based on variance according to existing studies \cite{ccarkouglu1998fairness,raman2021data} and accumulated utility acquired by each service provider $n$, shown as $ var\left(u_n\right)$. Similar to service provider-based fairness, the customer-based fairness in DVRP is defined based on variance and utility associated to each customer, shown as $var\left(u^c\right)$. For different case studies, there are differences on the formula to calculate $u_n$ and $u^c$ shown in the following subsections. In this study, we aim to balance between optimising two-sided fairness and utility.

\subsection{Ridesharing Systems and Fairness}

Ridesharing can be constructed as DVRPs as customers raise requests across time, to allow longer look-ahead time with consideration of optimising the total utility and fairness of the entire time duration \cite{anuar2021vehicle,abdullatif2020modelling,tan2021vehicle,wang2020decomposition}. Here, we consider a ride-hailing system, a sub-problem of ridesharing excluding the scenario of car-pooling. While existing studies exploit artificial intelligent algorithms such as reinforcement learning to optimise the total utility, it can bring fairness issues from both the side of the drivers such as unfair income distribution \cite{Lesmana2019,raman2021data,Xu2020,Nanda2020,Shi2021} and the side of the customers such as variant waiting time \cite{suhr2019two}. This can lead to issues such as insufficient amount of taxi drivers resulted from drivers with lower income quit the job \cite{shokoohyar2020impacts} and traffic congestion as customers tend to raise requests from the locations with shorter waiting period \cite{ortega2022algorithm}.



\subsection{Non-Compliance Capture and Fairness}

Non-compliance capture can be formulated as travelling officer problem (TOP) \cite{shao2019approximating} where non-compliances are raised from locations across time, which can be classified as DVRP. There are many existing solutions such as Ant Colony Algorithm, Genetic Algorithm, and Semi-Markov Reinforcement Learning \cite{shao2017traveling,qin2020solving,schmoll2021semi}. While the origin goal of TOP is to maximise the number of non-compliance caught, the works can lead to neglecting the importance of fairness in workload distribution among officers and capture rates in different areas \cite{ab2022survey,liljegren2021police,brazil2020unequal}. This can lead to issues such as officer demotivation and the psychology of fare evasion \cite{delbosc2019people}.


\subsection{Utility}

\textbf{Non-Compliance Capture} assume there are multiple officers that can be treated $N$ initialised at different locations. The officers patrol and issue parking tickets to the vehicles in non-compliance, and the output is the route plan for each officer. The cars parking at different parking bays are treated as $C$, where $s_c = null$ as we only focus on where the cars are parked and invoke parking non-compliance. The utility $B(n, c, t)$ gained by $n\in N$ from capture non-compliance invoked by $c\in C$ at time $t$ is defined as the parking fine obtained:

\begin{equation}
    B\left( n, c, t\right) =\begin{cases}
    1 & t^c_{s} \ \leqslant \ t\ \leqslant \ t^c_{e} \ \land \ l_{n} \ =\ d_c,\\
    0 & \text{ otherwise. }
    \end{cases}
    \label{utility_parking}
\end{equation} 

Thus, optimising total utility under this scenario is to maximise the total number of non-compliance captured by $N$ in a certain time duration $T$ shown as:

\begin{equation}
    \ \ \sum _{n\in N} u_n = \ \ \sum _{n\in N}\sum _{t=0}^{T}\sum _{c\in C} B\left( n,c,t\right)
    \label{total_utility_parking}
\end{equation}

\textbf{Ridesharing} assumes there are multiple drivers that can be treated $N$ initialised at different locations. The drivers are assigned to different requests raised by riders treated as $C$. We assume $t^c_e = \infty$ to simplify the scenario according to existing studies on improving fairness \cite{sun2022optimizing,suhr2019two,Shi2021}, which indicates the riders will wait until the request is picked up by $n\in N$. The utility $R(n, c, t)$ gained by $n\in N$ from being assigned to requests raised at $t$ by $c\in C$ is defined as:

\begin{equation}
    R( n,c,t) \ =\ Geo\left( d_{c} ,s_{c}\right) \ -Geo\left( s_{c} ,l_{n}\right)
    \label{utility_ridesharing}
\end{equation}
where $Geo$ is used to calculate Haversine distance \cite{van2012heavenly}. Thus, optimising total utility under this scenario is to maximise the total utility gained by $N$ in a certain time duration $T$ shown as:

\begin{equation}
    \ \ \sum _{n\in N} u_n = \ \ \sum _{n\in N}\sum _{t=0}^{T}\sum _{c\in C} R( n,c,t)
    \label{total_utility_ridesharing}
\end{equation}

\subsection{Service Provider-based Fairness}




\textbf{Service Provider-based Fairness in Non-Compliance Capture} is treated as the equal working load among officers, thus:

\begin{equation}
    \ \ var\left(u_n\right) = var\left(\sum _{t=0}^{T}\sum _{c\in C} B(n,c,t)\right)
    \label{service_provider_fairness_parking}
\end{equation}

\textbf{Service Provider-based Fairness in Ridesharing} is treated as the equal gain among drivers, thus:

\begin{equation}
    \ \ var\left(u_n\right) = var\left(\sum _{t=0}^{T}\sum _{c\in C} R(n,c,t)\right)
    \label{service_provider_fairness_ridesharing}
\end{equation}

\subsection{Customer-based Fairness}

In this study, we define customer-based fairness according to area-based fairness \cite{Brandao2020,Gollapudi2020,suhr2019two} by dividing customers into different spatial groups based on their starting locations $n_c^i$, where

\textbf{Customer-based Fairness in Non-Compliance Capture} is treated as the equal capture rate among different areas, defined as:

\begin{equation}
    \ \ var\left(u^c\right) = var\left( \ \ \sum _{n\in N}\sum _{t=0}^{T}\sum _{c\in g( C)} B(n,c,t)\right),
    \label{customer_fairness_parking}
\end{equation}

where $g(C)$ represents a list of parking non-compliance happened in the same area.

\textbf{Customer-based Fairness in Ridesharing} are treated as the equal mean waiting time among different areas, defined as:

\begin{equation}
     \begin{array}{l}
    \ \ var\left(u^c\right) = var\left( \ \ \frac{\sum _{t=0}^{T}\sum _{c\in g( C)} W(c,t)}{|g( C) |}\right) ,\ where\\
    \ \ \ \ \ W( c,t) \ =\ t\ -\ t^c_s,
    \end{array}
    \label{customer_fairness_ridesharing}
\end{equation}

where $g(C)$ represents a group of customers who raised requests in the same area. 

In this study, we aim to seek a balance among three objectives, optimising utility, improving service provider-based fairness, and improving customer-based fairness, in DVRPs.

\section{Methodology}
\begin{figure}[t]
\centering
\includegraphics[width=\linewidth]{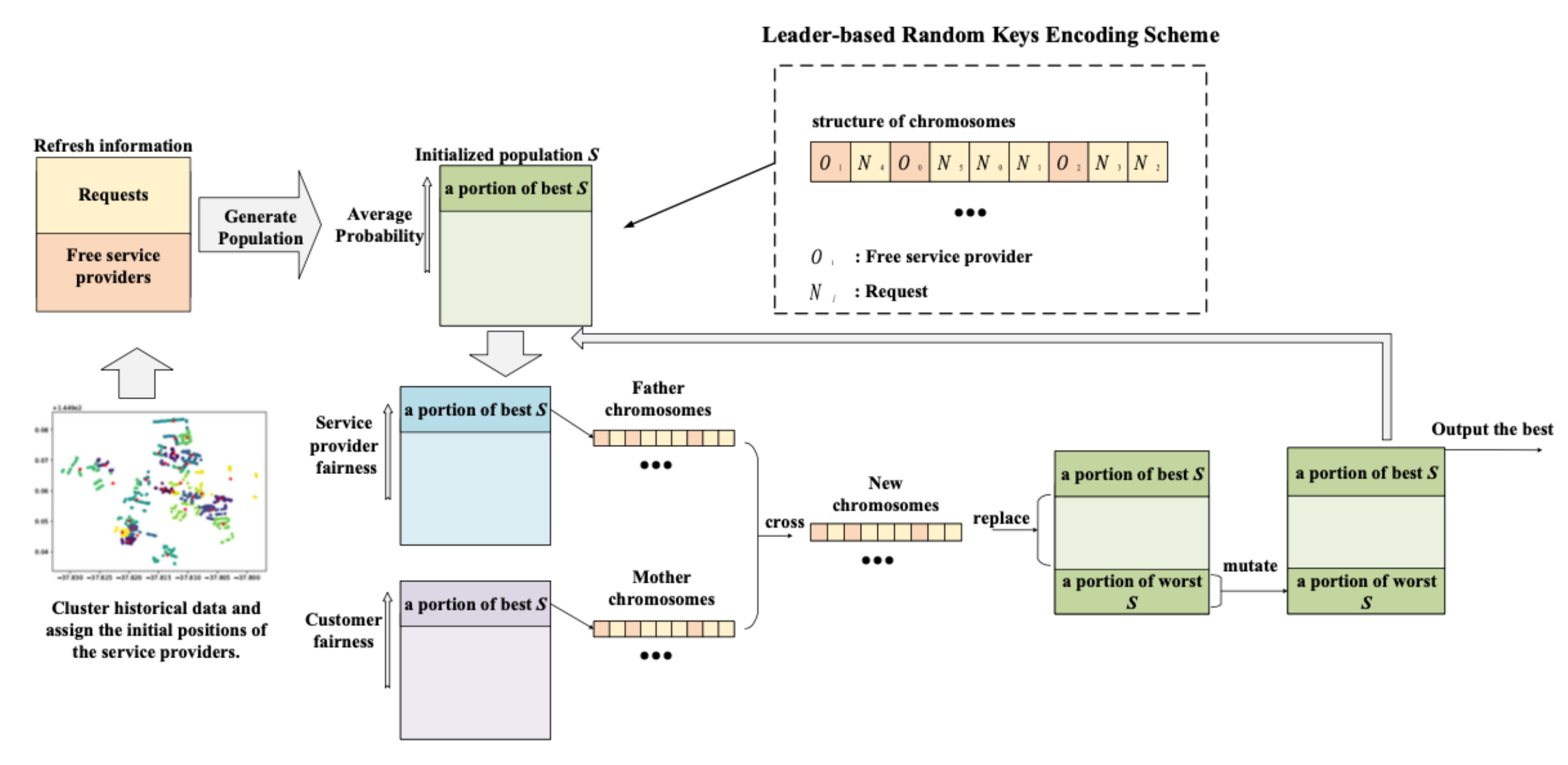}
\caption{Flowchart of our algorithm.}
\label{flow_chart_t}
\end{figure}

Existing methods for path optimisation problems can generally be divided into three categories: linear programming, heuristic algorithms, and reinforcement learning based approaches. As numbers of existing studies explore multi-objective optimisation based on evolutionary computation \cite{panwar2019many,pan2021practical,jain2013evolutionary,li2019predictive}, we exploit Genetic Algorithm (GA), which focuses on finding global optimal solutions through the iteration and mutation of elements, to propose Two-sided Fairness-aware Genetic Algorithm (2FairGA). While the original GA algorithm focuses on generating routing plans in a static environment, we propose 2FairGA that extends GA to a dynamic environment with multiple agents and balance between the three objectives. In 2FairGA, initial sampling focuses on initialising service providers to different locations in order to optimise two-sided fairness. Leader-based Random Keys Encoding Scheme (LERK) is used to incorporate the dynamic environment and multiple agents in DVRPs, and we modified two fitness functions in the GA algorithm to consider two-sided fairness while a local optimisation procedure is included in the modified GA to optimise total utility. In each iteration of 2FairGA, it generates an allocation and routing plan to allocate customers to different service providers and decide the sequence that each service provider should serve their allocated customers as shown in Figure~\ref{flow_chart_t}.

\subsection{Initial Sampling} 
Existing methods for optimisation initialise service providers in random locations or unoptimised, pre-defined origins. To optimise two-sided fairness, we propose to utilise a clustering algorithm to initialise service providers in different locations by sampling service providers to locations where similar travelling distances are required to travel to different customers. Additionally, to further enhance two-sided fairness, we propose to set up an additional constraint to balance the number of locations for potential customers around each service provider. To address the issues, we have utilised the constrained K-means clustering method to guarantee the minimum number of nodes in each cluster \cite{bradley2000constrained}. As shown in Eq.~\ref{CKMEANS} constrained K-means clustering includes an additional constraint aimed at guaranteeing the minimum number of nodes $\tau$ in each cluster $k$. 

\begin{equation}
\label{CKMEANS}
\begin{array}{cc}
\underset{T}{\operatorname{minimize}} & \sum_{i=1}^m \sum_{h=1}^k T_{i, h} \cdot\left(\frac{1}{2}\left\|x^i-C^{h, t}\right\|_2^2\right) \\
\text { subject to } & \sum_{k=1}^m T_{i, h} \geq \tau_h, h=1, \ldots, k \\
& \sum_{h=1}^k T_{i, h}=1, i=1, \ldots, m, \\
& T_{i, h} \geq 0, i=1, \ldots, m, h=1, \ldots, k .
\end{array}
\end{equation}
where $m$ represents the total count of data points. It's important to note that $T_{ih}$ equals 1 when data point $x_i$ is nearest to the centre $C_h$, and it's 0 in all other cases.

Based on the results of the constrained K-means clustering, 2FairGA distributes service providers to the centroids of the clusters. The number of service providers assigned to each centroid is proportional to the number of customers that tend to occur at the locations within the area.

\subsection{Leader-based Random Keys Encoding} 

To allow 2FairGA to incorporate temporal features under the two real-world scenarios, we utilise the Leader-based Random Keys Encoding Scheme (LERK) to generate temporary paths~\cite{qin2020solving}. It allows for the arrangement of 0 or more target locations for service providers. Service providers may not be assigned to target locations when the number of idle service providers exceeds the number of locations where the requests are raised by customers or when the service providers are too far away from all the existing requests. The scheme generates an initial solution by initialising random numbers and sorting them. The element between two service providers represents the target locations assigned to the former one. Thus, the first element in the sequence is forced to be an idle service provider as the leader.

\subsection{Genetic Algorithm}

To alleviate the conflict between two-sided fairness and utility optimisation and avoid the challenge of searching optimal solution from a non-convex function constructed by directly combining the three objectives, 2FairGA aims to balance the three objectives in different steps. To allow Genetic Algorithm (GA) to strike a balance between three objectives, we modify the two fitness functions in GA to consider service provider-based fairness and customer-based fairness accordingly, and we include an additional procedure in the original GA to apply local optimisation to optimise utility. 

Algorithm~\ref{FAIR-GA} shows the GA-related module (FairGA) in 2FairGA. 
\begin{algorithm}[!ht]
  \caption{FairGA}
  \label{FAIR-GA}
  \begin{algorithmic}[1]
    \Require  a set of requirements raised by customers \(C\)
    \Statex a set of service providers in free time \(M_{f}\)
    \Ensure
      The optimal solution \(S_{best}\)
      \State Initialize \(S\)
    	\While {\(Iteration<mIteration\)}
        \State Rank(\(S\), probability)
        \State Remain a portion \textit{elitistRate} of best \(S\) unchanged
        \For {\(i=0\); \(i<crossSize\); \(i++\)}
        \State Deep Copy \(S\) to get \(S_1\) and \(S_2\)
        \For{chromosome in \(S_1\)}
        \State Solution\(\leftarrow\)AssignProvider(provider-fairness)
        \EndFor
        \State Rank(\(S_1\),service provider)
  
       \For{chromosome in \(S_2\)}
        \State Solution \(\leftarrow\) AssignProvider(customer-fairness)
        \EndFor
        
        \State Rank(\(S_2\),customer)
        \State \(S_{mo} \leftarrow\) SelectCrossRate(\(S_1\)) 
        \State \(S_{fa} \leftarrow\) SelectCrossRate(\(S_2\))
        \State \(S_{child} \leftarrow\) Cross(\(S_{mo}\), \(S_{fa}\))
        \If {\(\text{uniform(0,1)}>localRate\)}
        \State Apply local optimization on \(S_{child}\)
        \EndIf
       \State \(S_i \in S \leftarrow S_{child}\)
        \EndFor
        \State Mutate(\(mutateRate, S\))
        \EndWhile
        \State Return \(S_{best}\)
  \end{algorithmic}
\end{algorithm}
\textbf{Rank(\(S\), parameter)}
Rank the population of chromosomes \(S\) in descending order of parameter. There are three types of parameters:
\begin{enumerate}[(1)]
    \item \textbf{probability}
\begin{equation}
    probability=\frac{\sum_n\sum_c p_n^c}{N_c},
\end{equation}
where \(p_n^c\) is the probability that the service provider is assigned to a request and gain utility. It represents the probability that the service providers arrives at the location before the customers
leave, which is only applicable in Non Compliance Capture. \(N_c\) is the number of requests raised at the current time.
    \item \textbf{Service Provider} This parameter represents service provider-based fairness, which can be calculated as Eq. \ref{service_provider_fairness_parking} or Eq. \ref{service_provider_fairness_ridesharing}.
    \item \textbf{customer} This parameter refers to customer-based fairness, defined as Eq. \ref{customer_fairness_parking} or Eq \ref{customer_fairness_ridesharing}.
\end{enumerate}
\textbf{AssignV(type)}
'type' represents the type of fairs. For each idle service provider, assign a candidate location to the service provider and then calculate either service provider-based or customer-based fairness. Move the location that improves fairness to the position closest to the service providers in the sequence. After all the idle service providers have been assigned a candidate location, output this assigned solution. \\
\textbf{SelectCrossRate(\(S\))}
Select a portion \(crossRate\) from top in \(S\). \\
\textbf{Cross(\(S_{mo}\), \(S_{fa}\))} employs single-point crossover. \\
\textbf{Mutate(\(mutateRate, S\))} employs swap mutation to immigrate a portion \textit{mutateRate} of worst \(S\). \\
\textbf{Local Optimisation} (line 21 in Algorithm 1) re-orders the sub-paths for each service provider by prioritise the requests with highest $p_n^c$, the length of the sub-paths is limited.

\begin{table}[]
  \centering
  \resizebox{0.4\textwidth}{!}{
    \begin{tabular}{l|l}
    \toprule
    \toprule
    Attribue & Description \\
    \midrule
    \midrule
    Area ID & The serial number of the area in the city \\
    Lat\& lon &  The location information of the parking bay\\
     Arrive Time & The time that a vehicle arrives at the parking bay\\
    Violation Time &The time that a vehicle started the violation \\
    Departure Time&The time that a vehicle leaves the parking bay \\
    Marker &  The signs placed on the side of parking bay\\
    \bottomrule
    \bottomrule
    \end{tabular}}%
    \caption{Attributes and descriptions used in the experiment.}
  \label{Attribute}%
\end{table}%

\begin{table}[]
  \centering
  \resizebox{0.47\textwidth}{!}{
    \begin{tabular}{l|l|l}
    \toprule
    \toprule
    Parameter&Value & Description \\
    \midrule
    \midrule
    elitistRate &0.2& Proportion of population for the reproduction operation \\
    crossRate &0.3&  Proportion of population for the crossover operation \\
     mutateRate & 0.2&Proportion of population for the immigration operation \\
    localRate& 0.5& Probability of performing local optimization\\
    populationSize &100 &The number of individuals in a population \\
    maxGen&300&The number of rounds in an iteration\\
    \bottomrule
    \bottomrule
    \end{tabular}}%
    \caption{Parameters used in the experiment in FairGA.}
  \label{parameter}%
\end{table}%

\begin{table*}[]
\resizebox{\textwidth}{!}{
\begin{tabular}{ccccccccccccc}
\hline
\multirow{2}{*}{Algorithm} & \multicolumn{3}{c}{Service Provider-based Fairness} & \multicolumn{3}{c}{Customer-based Fairness} & \multicolumn{3}{c}{Non-Compliance Captured} & \multicolumn{3}{c}{Total Distance} \\ \cline{2-13} 
 & \multicolumn{1}{c}{20} & \multicolumn{1}{c}{30} & 50 & \multicolumn{1}{c}{20} & \multicolumn{1}{c}{30} & 50 & \multicolumn{1}{c}{20} & \multicolumn{1}{c}{30} & 50 & \multicolumn{1}{c}{20} & \multicolumn{1}{c}{30} & 50 \\ \hline
2FairGA & \multicolumn{1}{c}{\textbf{466.68}} & \multicolumn{1}{c}{\textbf{857.19}} & \textbf{1340.31} & \multicolumn{1}{c}{0.0543} & \multicolumn{1}{c}{\textbf{0.0386}} & 0.0334 & \multicolumn{1}{c}{\textbf{1366.00}} & \multicolumn{1}{c}{\textbf{1614.00}} & 1772.00 & \multicolumn{1}{c}{\textbf{584217}} & \multicolumn{1}{c}{\textbf{683404}} & \textbf{531702} \\
GA & \multicolumn{1}{c}{1286.21} & \multicolumn{1}{c}{1884.35} & 2130.47 & \multicolumn{1}{c}{0.0603} & \multicolumn{1}{c}{0.0443} & 0.0313 & \multicolumn{1}{c}{1319.67} & \multicolumn{1}{c}{1591.33} & 1775.67 & \multicolumn{1}{c}{\textbf{608028}} & \multicolumn{1}{c}{703697} & 556240 \\
Cuckoo & \multicolumn{1}{c}{1261.71} & \multicolumn{1}{c}{1997.11} & 2028.304 & \multicolumn{1}{c}{0.0598} & \multicolumn{1}{c}{\textbf{0.0433}} & 0.0326 & \multicolumn{1}{c}{\textbf{1373.00}} & \multicolumn{1}{c}{\textbf{1647.67}} & \textbf{1782.00} & \multicolumn{1}{c}{637424} & \multicolumn{1}{c}{\textbf{698106}} & 552693 \\
PSO & \multicolumn{1}{c}{878.03} & \multicolumn{1}{c}{1162.87} & 2438.20 & \multicolumn{1}{c}{0.0632} & \multicolumn{1}{c}{0.0479} & 0.0436 & \multicolumn{1}{c}{1234.67} & \multicolumn{1}{c}{1456.33} & 1691.00 & \multicolumn{1}{c}{719074} & \multicolumn{1}{c}{1037881} & 932854 \\
Greedy & \multicolumn{1}{c}{\textbf{606.33}} & \multicolumn{1}{c}{\textbf{759.18}} & 2414.72 & \multicolumn{1}{c}{\textbf{0.0481}} & \multicolumn{1}{c}{0.0520} & \textbf{0.0261} & \multicolumn{1}{c}{426.00} & \multicolumn{1}{c}{692.00} & \textbf{1805.00} & \multicolumn{1}{c}{825079} & \multicolumn{1}{c}{1237119} & 541248 \\
Nearest & \multicolumn{1}{c}{653.83} & \multicolumn{1}{c}{865.65} & \textbf{1327.54} & \multicolumn{1}{c}{\textbf{0.0463}} & \multicolumn{1}{c}{0.0516} & \textbf{0.0281} & \multicolumn{1}{c}{621.00} & \multicolumn{1}{c}{658.00} & 1356.00 & \multicolumn{1}{c}{828201} & \multicolumn{1}{c}{35781575} & \textbf{125895} \\
Reassign & \multicolumn{1}{c}{724.63} & \multicolumn{1}{c}{887.12} & 1375.23 & \multicolumn{1}{c}{0.0571} & \multicolumn{1}{c}{0.0572} & 0.0779 & \multicolumn{1}{c}{579.38} & \multicolumn{1}{c}{721.42} & 1354.26 & \multicolumn{1}{c}{685612} & \multicolumn{1}{c}{1587315} & 545698 \\
Reassign(3) & \multicolumn{1}{c}{995.56} & \multicolumn{1}{c}{1212.27} & 1653.36 & \multicolumn{1}{c}{0.0642} & \multicolumn{1}{c}{0.0611} & 0.0815 & \multicolumn{1}{c}{954.38} & \multicolumn{1}{c}{1021.57} & 1369.68 & \multicolumn{1}{c}{725354} & \multicolumn{1}{c}{3531689} & 621569 \\
NAdap & \multicolumn{1}{c}{841.10} & \multicolumn{1}{c}{969.37} & 1432.44 & \multicolumn{1}{c}{0.0751} & \multicolumn{1}{c}{0.0799} & 0.0925 & \multicolumn{1}{c}{865.73} & \multicolumn{1}{c}{921.13} & 1425.60 & \multicolumn{1}{c}{625785} & \multicolumn{1}{c}{701275} & 625894 \\
NAdap(3) & \multicolumn{1}{c}{778.55} & \multicolumn{1}{c}{1128.33} & 1455.37 & \multicolumn{1}{c}{0.0659} & \multicolumn{1}{c}{0.0759} & 0.0965 & \multicolumn{1}{c}{925.71} & \multicolumn{1}{c}{1054.32} & 1568.50 & \multicolumn{1}{c}{665878} & \multicolumn{1}{c}{854542} & 795989 \\ \hline
\end{tabular}
}
\caption{Performance comparison of algorithms on the non-compliance synthetic dataset. The total utility is defined as non-compliance captured according to Eq.~\ref{utility_parking} and~\ref{total_utility_parking} with total distance as another evaluation metrics for total utility. The lower value for total distance represents a better output. For fairness, a lower value represents a fairer output according to Eq.~\ref{service_provider_fairness_parking} and~\ref{customer_fairness_parking}.}
\label{syn_results_table}
\end{table*}

\begin{table*}[]
\resizebox{\textwidth}{!}{
\begin{tabular}{ccccccccccccc} 
\hline
\multirow{2}{*}{Algorithm} & \multicolumn{3}{c}{Service Provider-based Fairness} & \multicolumn{3}{c}{Customer-based Fairness} & \multicolumn{3}{c}{Non-Compliance Captured} & \multicolumn{3}{c}{Total Distance} \\ 
\cline{2-13}
 & 20 & 30 & 50 & 20 & 30 & 50 & 20 & 30 & 50 & 20 & 30 & 50 \\ 
\hline
2FairGA & 1055.87 & 1602.34 & \textbf{1966.82} & 0.1586 & \textbf{0.1154} & 0.0738 & \textbf{1157.00} & \textbf{1363.67} & 1520.33 & \textbf{632298} & 715276 & 544725 \\ 
GA & 1704.85 & 2292.77 & 2628.11 & 0.2125 & 0.1758 & \textbf{0.0677} & \textbf{1165.00} & 1363.50 & 1560.50 & 638919 & 676780 & \textbf{509564} \\ 
Cuckoo & 1427.73 & 2786.12 & 2114.49 & 0.1948 & 0.1545 & \textbf{0.0646} & 1150.33 & \textbf{1409.00} & 1561.00 & 634931 & \textbf{626524} & \textbf{505260} \\ 
PSO & 1230.63 & 1516.30 & 2502.31 & 0.1948 & 0.1646 & 0.1541 & 1070.33 & 1222.00 & 1429.67 & 709510 & 975057 & 902318 \\ 
Greedy & \textbf{886.32} & \textbf{1089.38} & 1976.15 & 0.1458 & 0.1521 & 0.1538 & 352.00 & 475.00 & 979.00 & 829978 & 1220441 & 1621714 \\ 
Nearest & 1039.39 & \textbf{1131.24} & \textbf{1709.49} & \textbf{0.1448} & \textbf{0.1519} & 0.1526 & 359.00 & 496.00 & 998.00 & 834339 & 1216743 & 1609346 \\ 
Reassign & \textbf{1020.89} & 1520.64 & 2021.34 & 0.2128 & 0.2539 & 0.2013 & 1051.00 & 1228.00 & 1519.00 & 635176 & 694689 & 563814 \\ 
Reassign(3) & 1352.34 & 1428.32 & 1968.49 & 0.3087 & 0.3678 & 0.3024 & 954.00 & 1134.00 & 1494.00 & 638179 & 704590 & 572653 \\ 
NAdap & 1090.21 & 1385.31 & 1996.35 & \textbf{0.1415} & 0.2497 & 0.2971 & 1153.00 & 1360.00 & \textbf{1567.00} & \textbf{575805} & \textbf{633723} & 588061 \\ 
NAdap(3) & 1230.55 & 1685.31 & 2121.11 & 0.2468 & 0.2789 & 0.3561 & 1078.33 & 1286.33 & \textbf{1728.00} & 682587 & 725483 & 812441 \\
\hline
\end{tabular}
}
\caption{Performance comparison with baselines on Melbourne Parking dataset. The total utility is defined as non-compliance captured according to Eq.~\ref{utility_parking} and~\ref{total_utility_parking} with total distance as another evaluation metrics for total utility, and the lower value for total distance represents a better output. In regards to fairness, a lower value represents a fairer output according to Eq.~\ref{service_provider_fairness_parking} and~\ref{customer_fairness_parking}.}
\label{real_data_results_table}
\end{table*}

\section{Experiments}
We evaluate 2FairGA using three datasets: a non-compliance synthetic dataset, the Melbourne Parking dataset, and the New York City Taxi dataset. The evaluation examines the effectiveness of the proposed method against various baseline algorithms, focusing on metrics such as service provider-based fairness, customer-based fairness, and the total utility acquired by different service providers. An ablation study is also included to assess the impact of different components of the method and explore the correlation between the three objectives. The aim is to demonstrate the ability of the proposed method to effectively balance two-sided fairness and utility in different real-world scenarios formulated as DVRPs.

\subsection{Datasets}
\textbf{Non-Compliance Synthetic Dataset}
Our synthetic dataset emulates non-compliance in a square area using Poisson and exponential distributions~\cite{shao2017traveling}. The Poisson distribution is well-suited for representing the number of random events per unit of time, and vehicle stay time in parking lots adhere to an exponential distribution. We employ the Poisson distribution to generate the quantity of illegal parkings in each parking bay and the exponential distribution to determine the duration of each illegally parked vehicle.

\textbf{Melbourne Parking Dataset} The dataset is used for the case study of non-compliance capture. We acquired a dataset containing 2016 parking events in Melbourne, Australia, from the Open Data Platform~\cite{sensor}\cite{parking}. This dataset encompasses all parking events occurring within the on-street car parking bays of the Central Business District (CBD) over the course of one year. We primarily utilise data pertaining to six attributes detailed in Table \ref{Attribute}. We selected a day with the highest number of data records, February 8th as our test date, which had 1,657 instances of illegal parking.

\textbf{New York City Taxi Dataset} This dataset is used for the case study of ride-hailing. We acquired a dataset containing requests raised from different locations in 2016 in Mahattan, New York City (NYC), the United States. We selected a time period with the highest number of records, 6 pm 19th March as our test date, which  contains 10,000 instances of requests raised by different riders.

\subsection{Experimental Details}


\subsubsection{Assumptions}
To facilitate the evaluation of our algorithms, we make the following assumptions:

\begin{enumerate}
    \item As assumed in existing studies, service providers travel at a constant speed of 70 meters per minute \cite{shao2017traveling}.
    \item The status of service providers and requests raised by customers on the map are updated every minute.
    \item Service providers can only be given instructions when they are located at positions that correspond to nodes in a graph, which is constructed based on the real-world map.
    \item Each request only accept no more than one service provider.
\end{enumerate}

Parameters and descriptions are shown in Table \ref{parameter}.

\subsubsection{Baselines} We have selected multiple state-of-the-arat baselines to compare against.  They either focus on optimising utility or balancing fairness and utility that can be used in DVRP.  These include Greedy-Distance, Greedy-Probability, LERK-GA, LERK-PSO, LERK-CS \cite{qin2020solving}, Reassign \cite{Lesmana2019}, and NAdap \cite{Nanda2020}. The Reassign(3) and NAdap(3) are modified to consider the three objectives via $f\ =\ F_{provider} \ +\ ( 1-\lambda ) F_{customer}$ where $F_{provider}$ is defined based on Eq.~\ref{service_provider_fairness_parking} or Eq.~\ref{service_provider_fairness_ridesharing} and $F_{customer}$ is defined based on Eq.~\ref{customer_fairness_parking} or Eq.~\ref{customer_fairness_ridesharing}. As Reassign and NAdap are the two existing algorithms that consider single-sided fairness and there is not an existing study that consider two-sided fairness in a single model to the best of our knowledge, we include additional baselines that are modified to consider two-sided fairness based on Reassign and NAdap.

\subsection{Results and Analysis}

\begin{table}[!htbp]
\resizebox{\linewidth}{!}{
\begin{tabular}{cccc} 
\hline
Algorithm & Service Provider-based Fairness & Customer-based Fairness & Benefit \\ 
\hline
2FairGA & \textbf{6988.43} & \textbf{2.976759} & \textbf{2144904.33} \\ 
GA & 28235.96 & 10.6538 & -3481396.43 \\ 
Cuckoo & 17256.79 & 7.6528 & -4836859.24 \\ 
PSO & 20356.21 & 8.1354 & -4215940.31 \\ 
Greedy & 29532.17 & 11.8590 & 6325.23 \\ 
Nearest & 11052.35 & 7.2568 & -5324987.29 \\ 
Reassign & \textbf{8953.27} & 5.2594 & \textbf{2220767.95} \\
Reassign(3) & 10523.28 & 11.3598 & 7589.62 \\ 
NAdap & 9053.24 & \textbf{4.3029} & 1197325.86 \\ 
NAdap(3) & 13684.58 & 7.3255 & 1213593.33 \\
\hline
\end{tabular}
}
\caption{Performance comparison with baselines on New York City Taxi dataset. The total utility is defined as benefit according to Eq.~\ref{utility_ridesharing} and~\ref{total_utility_ridesharing}. In regards to fairness, a lower value represents a fairer output according to Eq.~\ref{service_provider_fairness_ridesharing} and~\ref{customer_fairness_ridesharing}.}
\label{real_data_results_2_table}
\end{table}

\begin{table*}[!htbp]
\centering
\resizebox{\textwidth}{!}{
\begin{tabular}{ccccccccccccc} 
\hline
\multirow{2}{*}{Algorithm} & \multicolumn{3}{c}{Service Provider-based Fairness} & \multicolumn{3}{c}{Customer-based Fairness} & \multicolumn{3}{c}{Illegal Vehicles Captured} & \multicolumn{3}{c}{Total Distance} \\ 
\cline{2-13}
 & 20 & 30 & 50 & 20 & 30 & 50 & 20 & 30 & 50 & 20 & 30 & 50 \\ 
 \hline
2FairGA & 1055.87 & 1602.34 & \textbf{1966.82} & \textbf{0.1586} & \textbf{0.1154} & \textbf{0.0738} & \textbf{1157.00} & \textbf{1363.67} & \textbf{1520.33} & \textbf{632298} & 715276 & \textbf{544725} \\ 
GAClusterProviderFair & \textbf{596.04} & \textbf{1060.85} & 1989.74 & 0.2266 & 0.1811 & \textbf{0.0880} & \textbf{1177.00} & \textbf{1372.00} & 1501.50 & \textbf{617897} & \textbf{712108} & 669931 \\ 
GAFair & 762.37 & 1454.00 & 2060.50 & 0.2039 & 0.1636 & 0.1247 & 1150.00 & 1330.50 & \textbf{1517.75} & 682742 & 824238 & \textbf{562772} \\ 
GAProviderFair & \textbf{715.57} & \textbf{938.61} & \textbf{1649.83} & 0.2137 & \textbf{0.1462} & \textbf{0.0880} & 1143.17 & 1356.00 & 1502.67 & 673369 & \textbf{700793} & 678571 \\ 
GACustomerFair & 961.11 & 1712.07 & 3132.53 & \textbf{0.1867} & 0.1613 & 0.1722 & 1101.33 & 1304.00 & 1439.00 & 727838 & 811692 & 774492 \\ 
GA(3) & 904.85 & 1192.77 & 2028.11 & 0.2125 & 0.1758 & 0.1777 & 1055.00 & 1363.50 & 1460.50 & 638919 & 806780 & 739564 \\
\hline
\end{tabular}
}
\caption{Ablation study using Melbourne Parking dataset with different modules removed from 2FairGA to test the necessity of different modules and the correlation between utility optimisation and two-sided fairness. }
\label{ablation_table}
\end{table*}

\begin{figure*}[!htbp]
    \centering
    \begin{subfigure}[c]{0.33\textwidth}
         \centering
         \includegraphics[width=\textwidth]{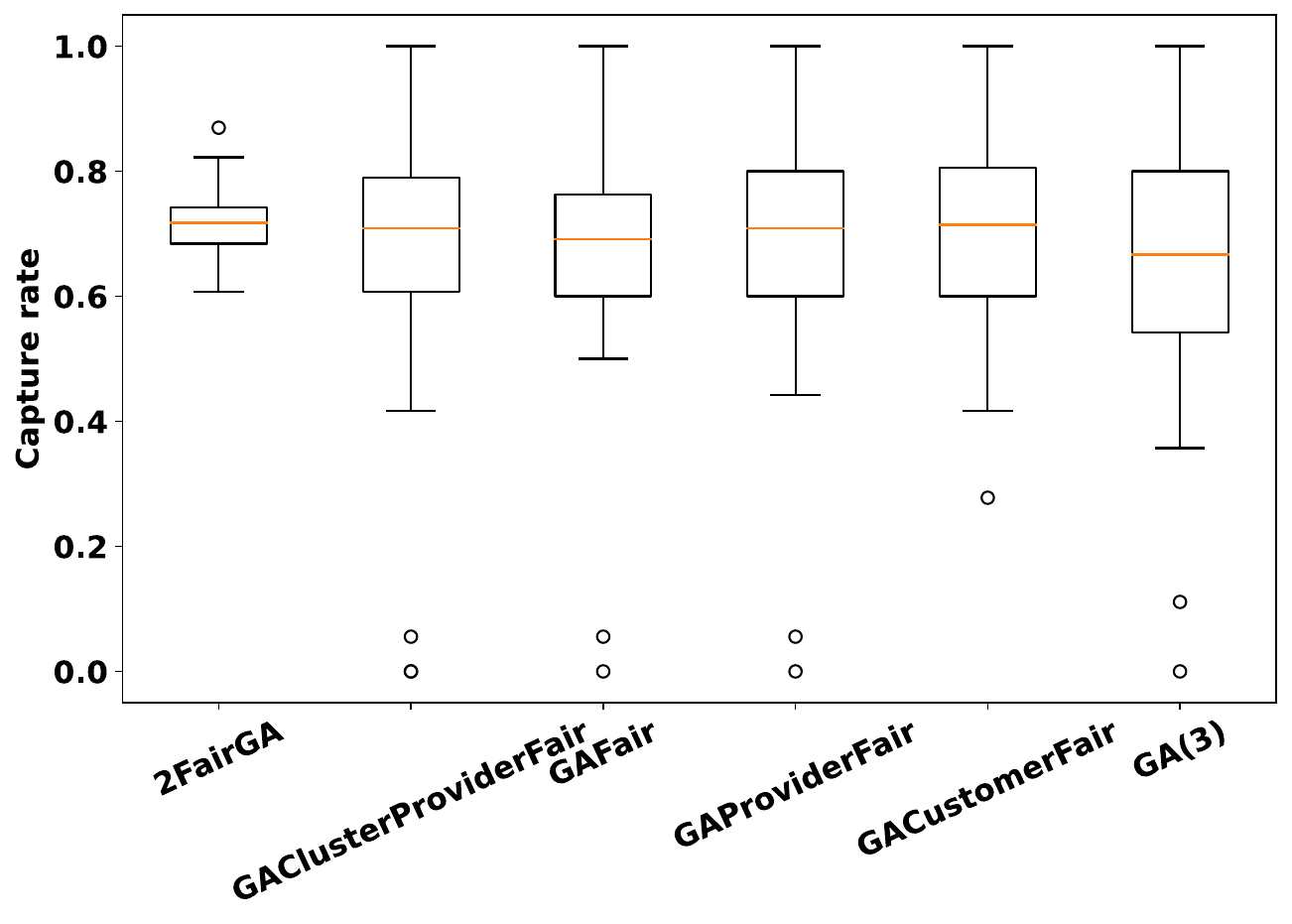}
         \caption{Capture rate with 20 service providers.}
         \label{fig:capture_rate_20}
     \end{subfigure}
    \begin{subfigure}[c]{0.33\textwidth}
         \centering
         \includegraphics[width=\textwidth]{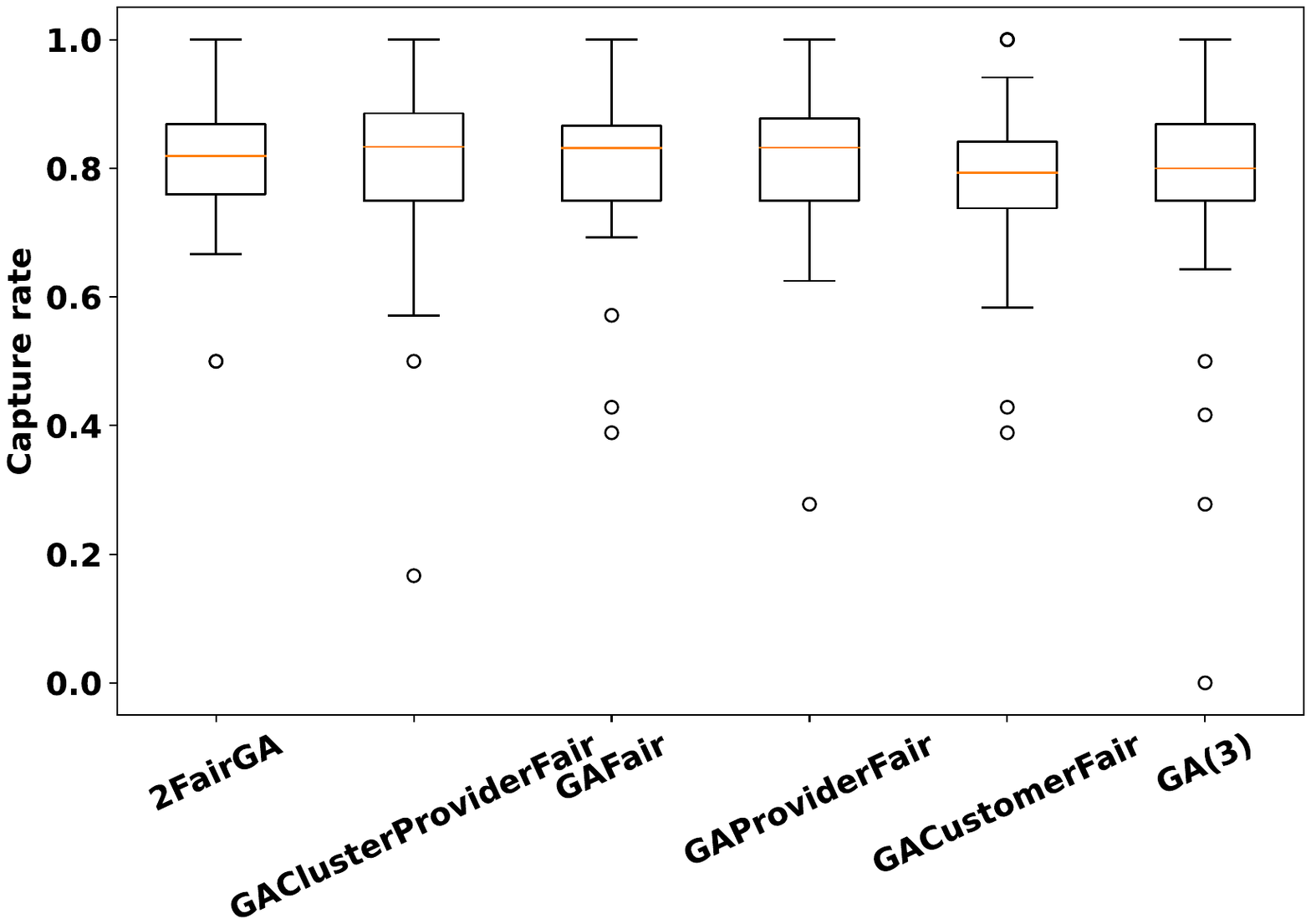}
         \caption{Capture rate with 30 service providers.}
         \label{fig:capture_rate_30}
     \end{subfigure}
    \begin{subfigure}[c]{0.33\textwidth}
         \centering
         \includegraphics[width=\textwidth]{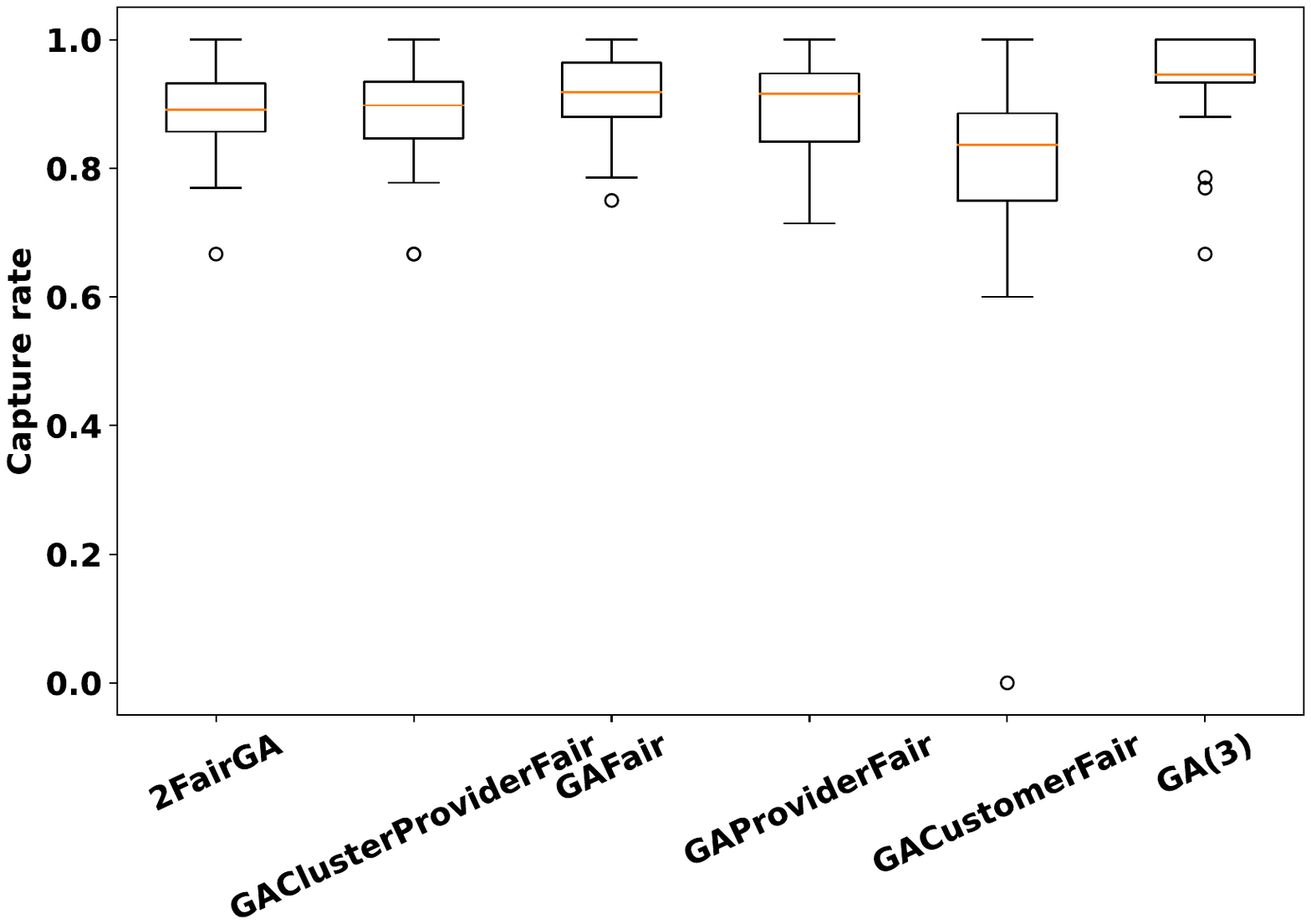}
         \caption{Capture rate with 50 service providers.}
         \label{fig:capture_rate_50}
     \end{subfigure}
     \begin{subfigure}[c]{0.33\textwidth}
         \centering
         \includegraphics[width=\textwidth]{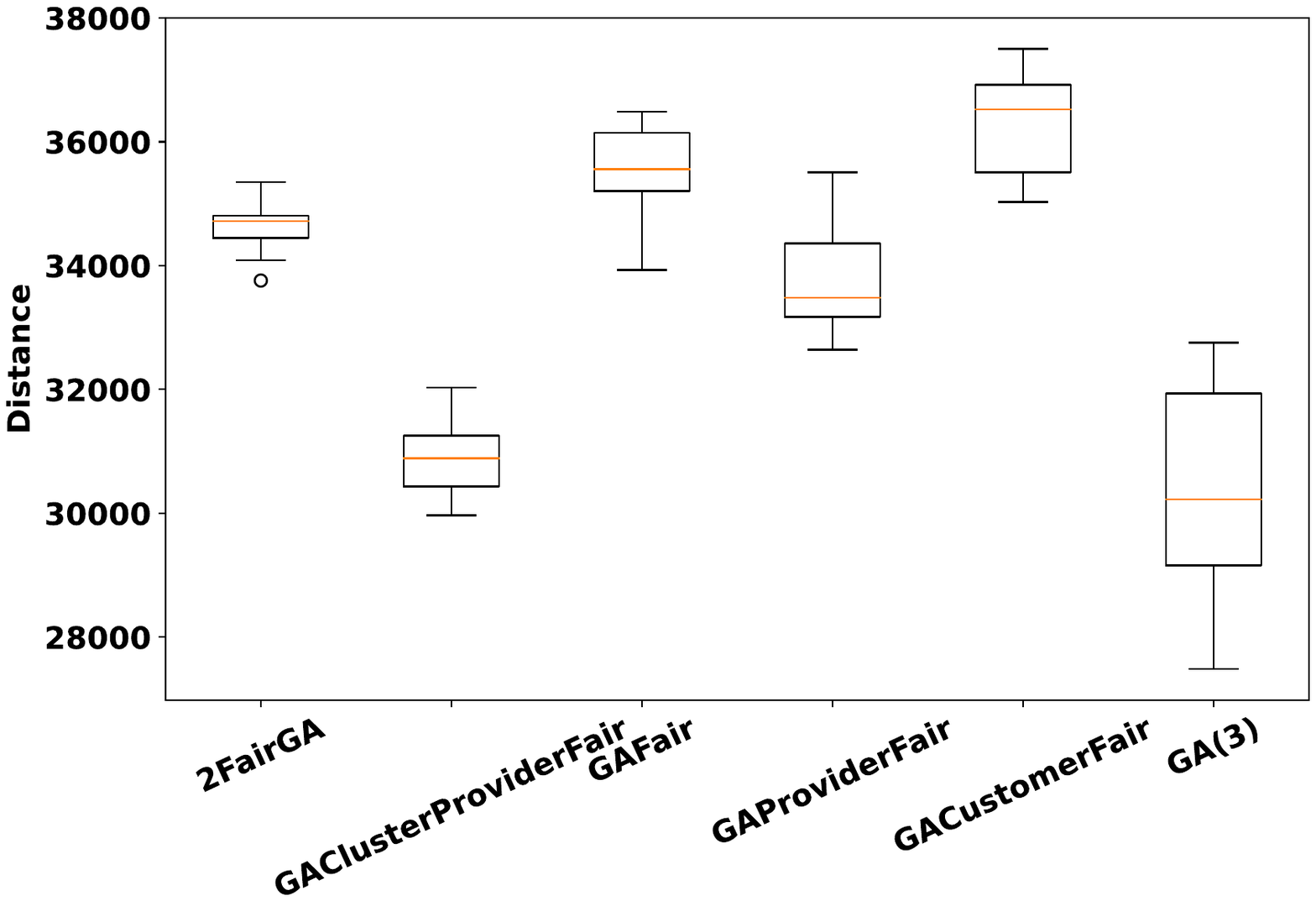}
         \caption{Total travel distance with 20 service providers.}
         \label{fig:Distance_20}
     \end{subfigure}
     \begin{subfigure}[c]{0.33\textwidth}
         \centering
         \includegraphics[width=\textwidth]{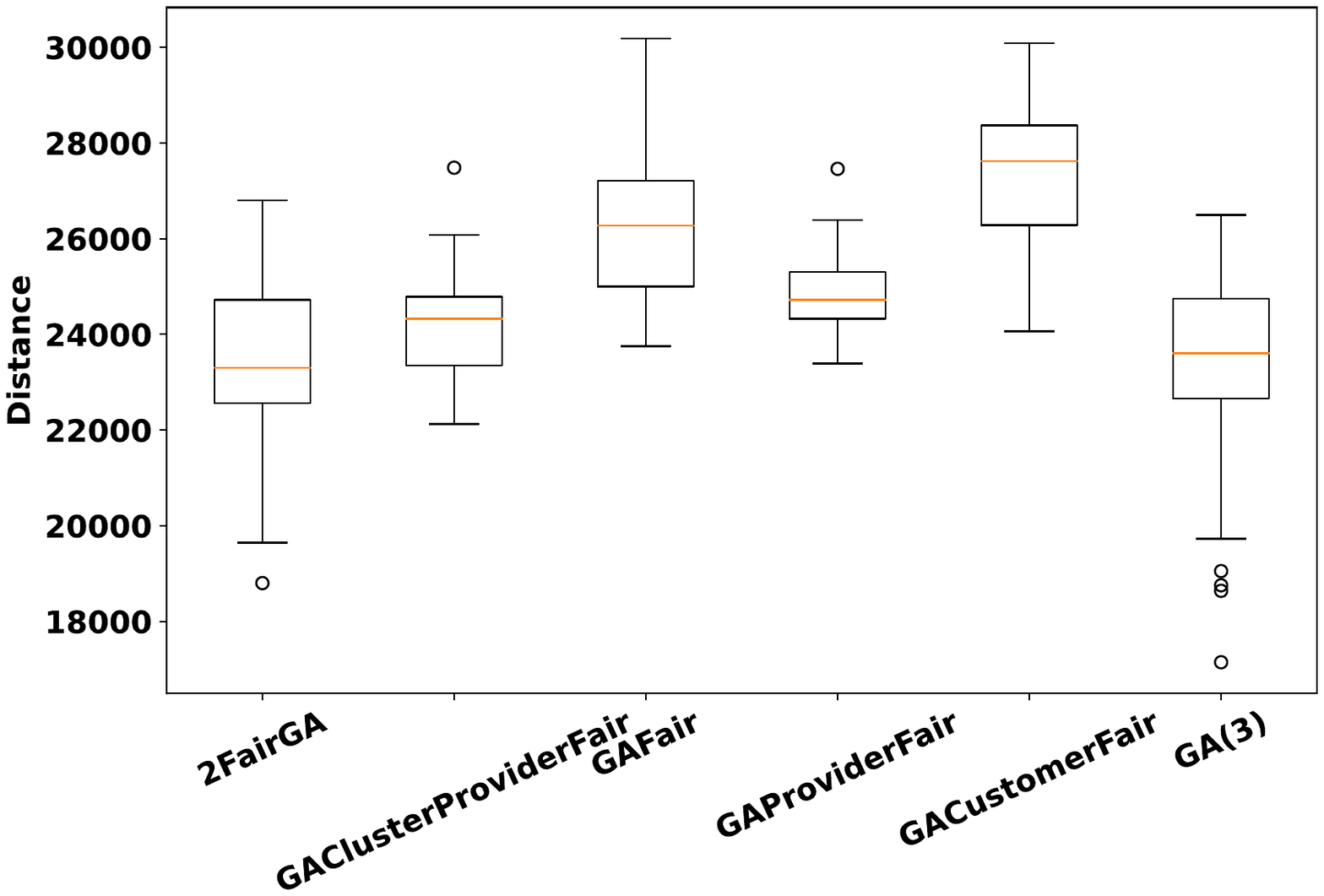}
         \caption{Total travel distance with 30 service providers.}
         \label{fig:Distance_30}
     \end{subfigure}
     \begin{subfigure}[c]{0.33\textwidth}
         \centering
         \includegraphics[width=\textwidth]{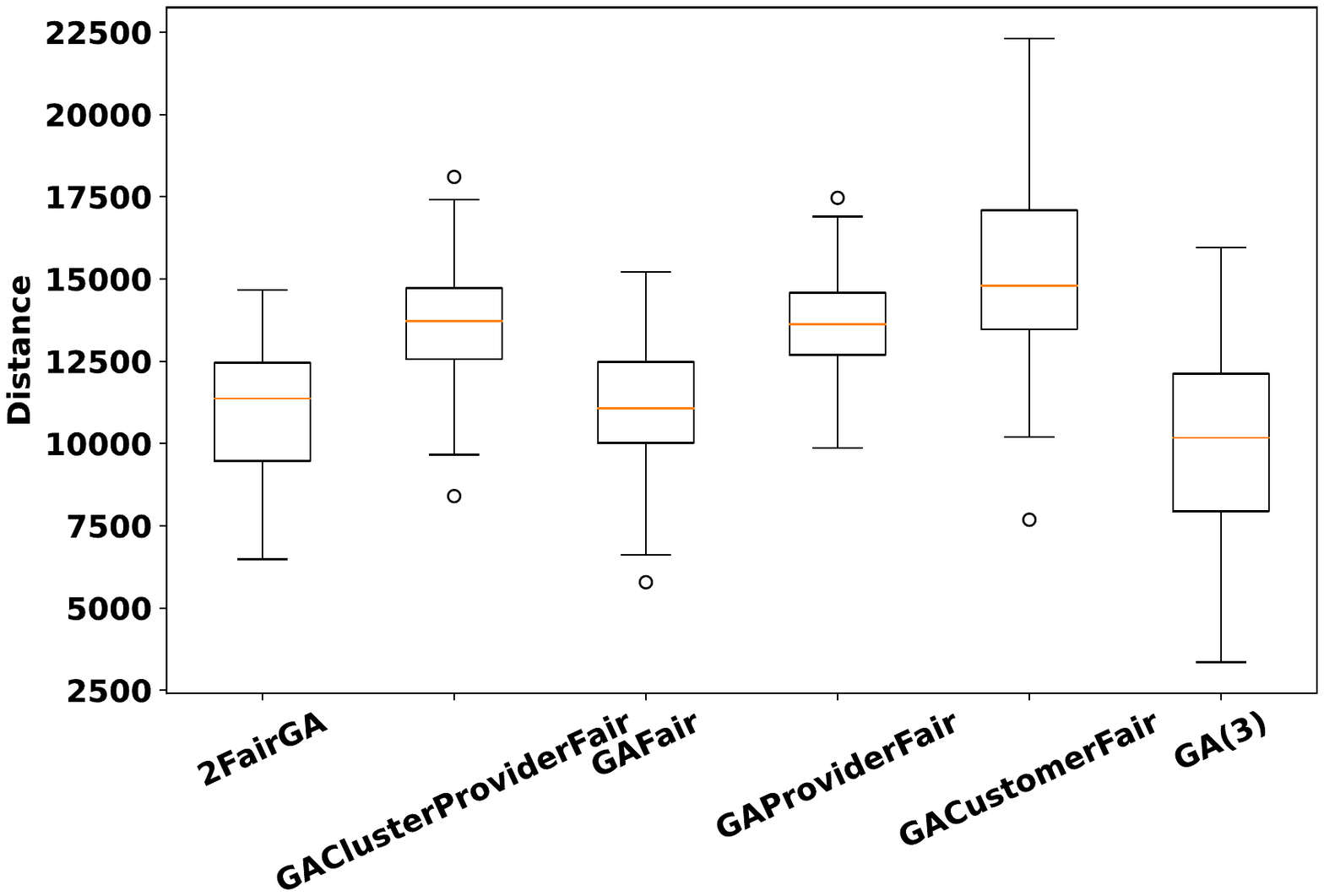}
         \caption{Total travel distance with 50 service providers.}
         \label{fig:Distance_50}
     \end{subfigure}
        \caption{Ablation study using Melbourne Parking dataset with different modules removed from 2FairGA to test the range of individual utility gained among service providers, which shows impact of different modules on range of fairness and utility.}
        \label{fig:abalation_study_boxplot}
\end{figure*}

Tables~\ref{syn_results_table},~\ref{real_data_results_table}, and~\ref{real_data_results_2_table} summarise the results of all the methods on three datasets. To better explore the performance of our proposed fairness-based optimisation approach, we incorporate the service provider-based fairness, customer-based fairness and clustering algorithm in the experiments as shown in Table~\ref{real_data_results_table} and Figure~\ref{fig:abalation_study_boxplot}. The best and the second best results are highlighted in boldface, and we analyse in detail in below.

\textbf{Comparison of Algorithms} 
Existing path optimisation algorithms either perform well on fairness or produce higher capture effectiveness at the expense of the other, while our proposed algorithm generally provides fair solutions and obtains high capture effectiveness at the same time.  Tables~\ref{syn_results_table}, ~\ref{real_data_results_table} and ~\ref{real_data_results_2_table} shows: 
\begin{enumerate}[(1)] 
\item Compared to other algorithms, 2FairGA have smaller service provider- and customer-based fairness while performing better for optimising utility and efficiency, more non-compliance captured and shorter total distance in the non-compliance capture scenario, and more total benefits in the ride-hailing scenario. It shows that 2FairGA can balance between multi-objectives because the proposed method considers different objectives in different steps. Additionally, 2FairGA shows significantly better performance on NYC taxi dataset compared to the baselines, which shows 2FairGA potentially generates good output. 
\item Under certain circumstances in non-compliance capture, Greedy and Nearest perform better on fairness, but their capture effectiveness is much lower than others against the original goal of the scenario. It is due to the characteristics of the datasets used for the non-compliance capture. There is a noticeable homogeneity in the spatial distribution of non-compliance locations, with a proximate quantity of infractions observed in the vicinity of each designated service providers. Under the ride-hailing scenario, by using the NYC taxi dataset, the two methods cannot provide good solutions for either of the objectives.

\item GA, Cuckoo and PSO have higher capture effectiveness than our approach. However, these algorithms gain higher scores on fairness metric, meaning that the solution provided by them is unfair because fairness are not considered in these baselines.
\end{enumerate}

Tables~\ref{syn_results_table},~\ref{real_data_results_table}, and~\ref{real_data_results_2_table} show that our algorithm is suitable for different scenarios. NYC Taxi dataset is the largest one among the datasets used, where the execution time for 2FairGA is around 10 hours.

\textbf{Ablation Study} 
To assess the individual contribution of each module within our proposed framework, we undertake a comprehensive ablation study on the Melbourne Parking dataset. We list the different combinations of each individual module as follows:

\begin{enumerate}
    \item 2FairGA: The proposed method in this study. 
    \item GAClusterProviderFair: excluding customer-based fairness. 
    \item GAFair: 2FairGA excluding initial sampling. 
    \item GAProviderFair: 2FairGA excluding initial sampling and customer-based fairness.
    \item GACustomerFair: 2FairGA excluding initial sampling and service provider-based fairness. 
    \item GA(3): Genetic Algorithm with fitness function designed by directly combine the three objectives ($U-F_{customer}-F_{provider}$, where $U$, $F_{provider}$, $F_{customer}$ are defined based on Eq.~\ref{utility_parking},~\ref{service_provider_fairness_parking},~\ref{customer_fairness_parking} accordingly).
\end{enumerate}

To explore the distribution of individual utility acquired by different service providers to assess fairness, we further visualise the experimental results of the ablation study using a box plot. As depicted in Figure~\ref{fig:abalation_study_boxplot}, our analysis yields the following observations:

\begin{enumerate}[(1)]
\item GACustomerFair, which focusing on customer-based fairness, exhibits a lower capture rate and a higher total distance compared to other ablation algorithms. Additionally, its performance is more variable.
\item Regarding capture rate, GAClusterFair displays more concentrated data when the number of service providers is equal to or greater than the number of areas. However, the range of acquired utility by different service providers becomes larger when the number of service providers is less than the number of areas, compared to GAFair.
\item Compared with only including service provider-based fairness (GAClusterProviderFair), 2FairGA, the proposed method which includes two-sided fairness, does not produce significantly more variant results in terms of the range of utility acquired by different service providers. It shows service provider-based fairness and customer-based fairness are slightly negative correlated.  
\end{enumerate}

From Table~\ref{ablation_table}, we draw the following conclusions:

\begin{enumerate}[(1)]
\item GACustomerFair, which emphasising customer-based fairness, exhibits better results in customer-based fairness metrics. 
\item GAClusterProviderFair and GAProviderFair perform better in terms of service provider-based fairness metrics. 
\item GAClusterProviderFair and GAProviderFair outperform GACustomerFair. GAClusterProviderFair captures more illegally parked vehicles and covers a shorter total distance. 
\item The integration of clustering algorithms results in fairer solutions and higher capture effectiveness compared to algorithms without clustering. Although GAFair has lower service provider-based fairness scores, 2FairGA displays better customer-based fairness and captures more non-compliance with shorter total distances, which shows the necessity of initial sampling in regards to different objectives
\item GA(3) does not output a good solution as the scale of the three objectives are significantly different. The outputs in terms of service provider-based fairness is better compare to the customer-based fairness and utility. The reason is that it puts a high weight on optimising service provider-based fairness as it has higher upper bound compare to the other two objectives. 
\end{enumerate}

\section{Conclusion}
This paper proposes a novel approach to solving the dynamic vehicle routing problem that considers two-sided fairness and utility optimisation. We designed two different fairness metrics, service provider and customer-based fairness, and integrated them with a Genetic Algorithm. We also highlighted the importance of initial sampling of service providers and used clustering to determine the starting location of each service provider, which has a significant impact on fairness and utility optimisation. Extensive experiments demonstrate that our proposed approach effectively balances fairness and utility optimisation, outperforming existing methods either focusing on only utility optimisation or considering fairness.

\begin{acks}
We acknowledge the support of the Australian Research Council (ARC) Centre of Excellence for Automated Decision-Making and Society (ADM+S) (CE200100005). This research was partially supported by NVIDIA Academic Hardware Grant Program. This research/project was undertaken with the assistance of computing resources from RACE (RMIT AWS Cloud Supercomputing).
\end{acks}

\bibliographystyle{ACM-Reference-Format}
\bibliography{gecco.bib}

\end{document}